# Real-time classification of EEG signals using Machine Learning deployment

**Swati CHOWDHURI[1*], Satadip SAHA[1], Samadrita KARMAKAR[1], Ankur CHANDA[2]**

[1*] Department of Electrical and Electronics Engineering, Institute of Engineering & Management, University of Engineering & Management (UEM), Kolkata, India

[2] Department of Computer Science and Engineering (Artificial Intelligence and Machine Learning), Institute of Engineering & Management, University of Engineering & Management (UEM), Kolkata, India

swati.chowdhuri@iemcal.com[*], satadipsaha3@gmail.com, bonnythemaa@gmail.com, ankurchanda198@gmail.com

**\*Corresponding Author:** Swati CHOWDHURI
swati.chowdhuri@iemcal.com[*]

**Abstract:** The prevailing educational methods predominantly rely on traditional classroom instruction or online delivery, often limiting the teachers' ability to engage effectively with all the students simultaneously. A more intrinsic method of evaluating student attentiveness during lectures can enable the educators to tailor the course materials and their teaching styles in order to better meet the students' needs. The aim of this paper is to enhance teaching quality in real time, thereby fostering a higher student engagement in the classroom activities. By monitoring the students' electroencephalography (EEG) signals and employing machine learning algorithms, this study proposes a comprehensive solution for addressing this challenge. Machine learning has emerged as a powerful tool for simplifying the analysis of complex variables, enabling the effective assessment of the students' concentration levels based on specific parameters. However, the real-time impact of machine learning models necessitates a careful consideration as their deployment is concerned. This study proposes a machine learning-based approach for predicting the level of students' comprehension with regard to a certain topic. A browser interface was introduced that accesses the values of the system's parameters to determine a student's level of concentration on a chosen topic. The deployment of the proposed system made it necessary to address the real-time challenges faced by the students, consider the system's cost, and establish trust in its efficacy. This paper presents the efforts made for approaching this pertinent issue through the implementation of innovative technologies and provides a framework for addressing key considerations for future research directions.

## 1. Introduction

The electroencephalogram (EEG) is a recording of the brain's electrical activity, captured through electrodes placed on the scalp. These electrodes detect and measure neural impulses between synapses as neurons communicate. EEG signals offer valuable insights into brain function, used in neuroscience, medicine and research for diagnosing and monitoring conditions like epilepsy, sleep disorders and brain injuries. Signal patterns and frequencies indicate brain activity states such as wakefulness, sleep stages, seizures, cognitive processes and abnormalities.

Machine learning can accurately measure EEG parameters like alpha, beta, theta and gamma waves values. Penolazzi et al. found that delta signals increase during linguistic acquisition, while high theta values indicate active concentration. Maximum alpha values suggest a state of rest or relaxation. These signals are computationally obtained using suitable algorithms, including machine learning models trained to recognize patterns and make predictions from given data.

Training a machine learning model involves dataset collection, handling missing values, scaling features, and dividing data into training and testing sets. An appropriate model is selected based on the problem type, and features are engineered or selected to improve pattern recognition. The model is trained iteratively to minimize errors and learn patterns, then its accuracy and generalization ability are tested. Hyperparameter tuning optimizes model performance before deploying the model for real-world applications. Implementing EEG devices in real-time learning environments aims to revolutionize education delivery. Privacy concerns and discomfort related to neural activity monitoring are valid, but the aim of this study is to address these issues while showing consideration for the study participants. In this study, EEG signals from students watching educational videos in a controlled setting were recorded and analysed in real time. The participants were informed and consented to data usage for the study's purposes.

The remainder of this paper is structured as follows: Section 2 elaborates on the scientific literature that was referred to during the development of this project. Section 3 provides a detailed description of the procedure followed during the data collection as well as evaluation and computation stages. In section 4, the process of data collection itself has been expanded upon, with both the quantitative and qualitative approaches being explained. Section 5 speaks of the feature extraction process that the data went through via 3 different classifiers. Next, the experiment itself has been elaborated on. The various study materials used, the 3 classifiers used along with the sensors and their application have been touched upon. The different types of signals received by the EEG sensors have then been operated on by the classifiers to plot the relevant correlation matrix. Finally, the results have been analysed showing both the individual performances of the classifiers as well as their performance as an ensemble model. The next section concludes the paper and brings to light the future prospects of this field of study.

## 2. Literature review

This paper explores the detection and evaluation of student focus and engagement with study materials using electroencephalography (EEG), a non-invasive brain imaging technique (Thibault et al., 2015). Various factors impacting data collection were considered. For instance, Beebe et al. (2010) demonstrated through EEG monitoring that adolescents experiencing sleep deprivation exhibited impaired learning and lack of attention in a simulated classroom. Additionally, mental fatigue significantly influences EEG fluctuations, with higher fatigue resulting in more chaotic outputs (Zhang et al., 2007).

Support vector machine (SVM) classifiers were employed to navigate the complex multivariate relationships typical of the EEG data (Cuingnet et al., 2010), facilitating the interpretation of plots through spatial regularization. SVMs are advantageous in managing unbalanced datasets (Tang et al., 2009). Bilucaglia et al. (2021) compared SVM, kNN, and LDA classifiers for analyzing EEG signals from emotion-related brain activity, concluding that SVMs provide the highest accuracy. Mohsin et al. (2022) proposed the use of the SMOTE algorithm to address class imbalance issues, finding that Random Forest and Naive Bayes also provide a commendable accuracy and sensitivity in data prediction (Chowdhuri et al., 2022). Alpha oscillations, recorded in the brain activity frequency band of 8-12 Hz after band-pass filtering, significantly decrease with activities involving thinking, working memory, and attention (Aranibar & Pfurtscheller, 1978; Becker et al., 2018). The work by Sun et al. (2014) explains that a stimulus on a neuron causes sodium ions to enter, raising the voltage which results in an "action potential" or electrical discharge. The EEG signal faces challenges due to the various filters it passes through, including the meninges, skull, and scalp. Various artifacts like heartbeats and eye blinking can also interfere with the EEG signals. A thorough signal processing and analysis are required for research utility (Miranda & Brouse et al., 2005). EEG signals cannot be influenced by untrained individuals, making them an appropriate measure of attentiveness (Costantini et al., 2010).

Penolazzi et al. (2008) used EEG delta band to examine word processing activity in children with dyslexia and control participants, validating EEG monitoring as a measure of comprehension. Seshadri et al. (2022) confirmed these findings in a brain network analysis during arithmetic problem-solving. Research shows that medial-frontal oscillations in the beta wave frequency range are associated with positive experiences, while lower frequencies relate to negative experiences (Boksem & Smidts, 2015). It is crucial to combine predictions from a "weak" algorithm during model training to minimize errors and bias (Opitz & Maclin, 1999). Proper sampling rates, above the Nyquist frequency, are necessary to avoid signal distortion (Swartz, 1998). Issues in EEG data collection, such as reference location and the number of electrodes, were addressed using an Emotiv Sensor (Srinivasan, 2012). Preparing the application site to minimize impedance is essential (Hari et al., 2023). Applications of EEG are vast, including neurofeedback for reducing epileptic seizures (Tan et al., 2009) and monitoring the progression of neurological disorders (Noachtar et al., 1999). This study aims to prove that monitoring EEG signals to evaluate student attention can help bridge the gap in engagement assessment between face-to-face and online education, enhancing the overall learning experience (Qi et al., 2023). Solutions in a similar vein, introduced by Huster et al. (2014), specifically considers the relevance of experimental and clinical trials for the enhancement of neurofeedback technology that relies on recordings of electroencephalograms. To facilitate the daily life of children suffering from ASD an innovative game has been designed that includes social interactions and provides neural and body-based feedback that corresponds to the trained signals that have been reinforced through this game (Friedrich et al., 2014).

## 3. Description

This paper evaluates the learning progression of a group of student participants by assessing their understanding of selected topics. These topics were chosen from a range of subjects including Mathematics, English, Physics, Chemistry, Biology, Psychology, Geography, History, and Sociology, tailored to each student's field of interest. The participants were briefed on the protocols to be followed during the examination, which involved wearing an EEG sensor on their scalp while studying the provided material without interruption. The EEG data, collected using Emotiv sensors, recorded the generated brain waves to evaluate each student's attentiveness during the learning process.

The Emotiv sensors captured frequency levels across various ranges, which were manipulated in real time to provide insights into the students' performance. The parameters of the EEG device were defined using a classification method for training a machine learning model. In this study a highly portable mobile sensor device was utilized, aiming to facilitate its use within classroom settings (Chowdhuri et al., 2024). The wireless sensors detected signals and transmitted them to hardware devices. The device used in this study featured a semi-dry sensor with two electrodes positioned in the parietal and occipital regions of the subject's head.

A SVM classifier was employed to classify the analysed data. By applying standard scaling, we centered the data by removing the mean from each feature and it was scaled to achieve unit variance. This process produced a linear set of values, which were instrumental in determining the mean and standard deviation. These values were crucial for achieving accuracy in training the model. Once the model is trained, its performance is validated using a different set of real values to check its predictive accuracy.

The prediction model used binary labels, where 0 indicated that the student had learned the topic and 1 indicated that the student had not. These predictions were implemented in a web application which was developed using Streamlit. Linear regression plots of the data were generated, illustrating predefined labels in relation to the total number of observations. This method provided a clear visualization of the learning progression and model accuracy, highlighting the effectiveness of using EEG data to monitor and evaluate the level of the students` engagement and comprehension.

## 4. Data collection

Data collection for this experiment was meticulously conducted to minimize observational errors. Participants were selected from a diverse pool of student applicants, representing various fields of study and economic backgrounds, thereby ensuring unbiased data. Both qualitative and quantitative data was collected. Quantitative data, which is numerical and computable, contrasts with qualitative data, which is descriptive and non-numerical, typically expressed in words (Taherdoost, 2021).

The Emotiv sensor was placed on the participants' heads, and they were instructed to study the provided materials with the purpose of learning the subject matter as it is shown in Figure 1.

The sensor recorded the frequency of electrical waves generated by neural activity during the study sessions, which constituted the quantitative data. This data was subsequently classified using various mathematical classifiers, as detailed later in this paper.

In parallel, qualitative data was gathered through observations of facial expressions indicating concentration, engagement, confusion, comprehension, and other relevant cues, as well as body movements and fidgeting hands. Collecting this qualitative data was essential, as it enabled a broader understanding of human behaviour, which could be correlated with the quantitative data from the sensor. This correlation is vital for assessing the accuracy of the sensor and the overall effectiveness and utility of the experiment in reaching a comprehensive conclusion.

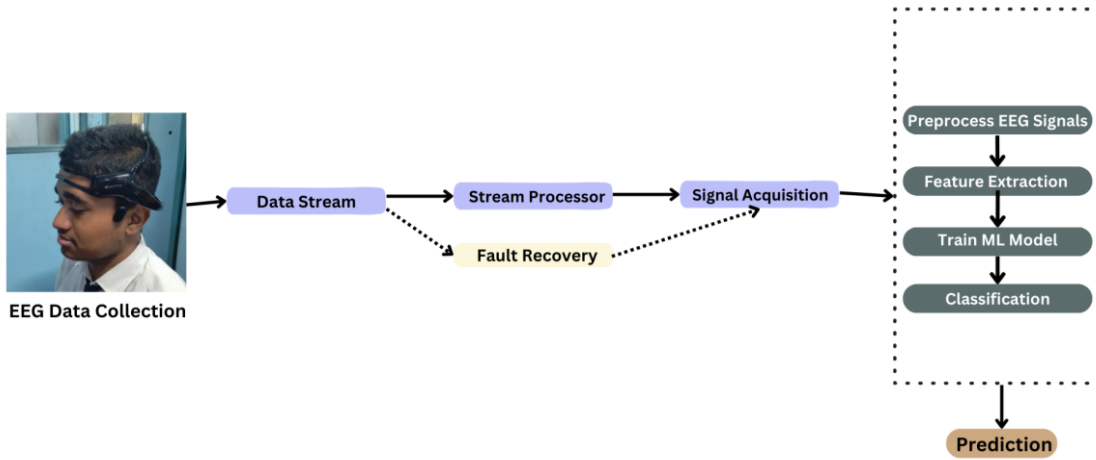

**Figure 1.** EEG signal analysis general steps

## 5. Feature extraction and EEG classification

Individual students exhibited different levels of attentiveness when presented with the same content, causing their EEG signals to fluctuate accordingly. This study aims to recognize these variations in EEG signals during focused learning under controlled conditions. In a supervised learning setting, the impact of the experiments on participants was determined and analysed. EEG signals were collected using Emotiv sensors, then filtered, processed, and prepared for analysis. The processed signal features were categorized into two categories: a predefined "attentive" label and a user-defined label, serving as the training set for a SVM classifier. After optimizing various classifier parameters, these classifiers were utilized to identify and record participant data. This data can assist educators in assessing students' attention levels.

### 5.1. Feature extraction

The EEG data was recorded at a sampling rate of 128 Hz with a 16-bit quantization level. To simplify the processing method, the recorded EEG data was passed through a low-pass filter with a cutoff frequency of 50 Hz. A fast fourier transform (FFT) transformed a segment of 128 sampling points into the frequency domain. Assuming that a function F(n)(n = 1, 2,….. 256) is the FFT result of a segment, the PSD (Power Spectral Density) can be expressed as:

$$P(n) = \frac{F(n) F \times (n)}{N} \qquad (1)$$

where F*(n) is the conjugate of F(n) and N= 256 sample points.

There are five EEG frequency bands, namely alpha, beta, delta, theta and gamma (Swartz, 1998) has identified that the frequency bands are most commonly associated with the human brain. The primary 4 energy bands take on different values according to the waveband distribution which leads to the following 4 features:

$$E_\alpha = \sum_{freq=8}^{13} P_{freq} \tag{2}$$

$$E_\beta = \sum_{freq=14}^{30} P_{freq} \tag{3}$$

$$E_\theta = \sum_{freq=4}^{7} P_{freq} \tag{4}$$

$$E_\delta = \sum_{freq=0.5}^{3} P_{freq} \tag{5}$$

where P is the value of energy for the given frequencies *freq*. Additionally, a previous work by Chowdhuri et al. (2024) has identified interconnections between α and β activities. The α activity indicates the brain being in a state of relaxation, while the β activity is associated with stimulation and alertness. In this study, five features related to these activities were extracted for classification purposes. The EEG signals were manually categorised as being related to the "attentive" and "unattentive" groups. These classifications were used to determine the accuracy rate for the classifier after completing the related calculations and classifications.

## 6. The experiment

First, the aim was to calculate the probability whether a given session was confusing for the students using Naive Bayes classifiers. This family of algorithms operates on the principle that each classified element is independent of the others. For this study, the analysed data was divided into two parts: a feature matrix encompassing the entire dataset, and vectors related to dependent features such as subject ID, video ID, attention, meditation, raw EEG signals, alpha, beta, gamma, delta and theta waves.

This study involved nine subjects who had no prior experience with EEG-related training. To clearly evaluate the states of attentiveness through EEG signals during learning, standard English materials were used. The materials covered various subjects including English, Mathematics, Physics, Chemistry, and Biology. Subsequently, different questions were asked to evaluate their learning process (questions included those requiring viewing and selecting images, multiple-choice questions, and selecting answers based on illustrations). This was done to thoroughly verify if the learners could concentrate during the experiment.

Before the official commencement of the experiment, students wore the sensors for at least two minutes to familiarize themselves with the sensation of wearing these and thus prevent any discomfort which could affect the accuracy of the results. The speaker volume was adjusted to ensure an optimal auditory environment for the learners. Predefined labels were manipulated to determine the students' level of confusion in understanding the provided study materials. The EEG equipment emitted various signals such as alpha, beta, theta waves, etc. Videos were presented randomly to prevent anticipation of the confusion level that could be generated by the upcoming videos. The students were given a minute to clear their minds before continuing to watch the instructional videos to maximize their learning capacity. After each session, students evaluated their level of confusion on a scale of 1 to 10, where 1 represented the lowest level of confusion and 10 the highest. Additionally, four observers rated the subjects' body language and expressions on a similar scale. Two-minute videos were recorded and any playback issues were noted. If a student was deemed not ready, the first and last 30 seconds were excluded, and only the 60 seconds in the middle of the video were analysed. Each student wore a wireless single-channel EEG detection headset for measuring brain activity across the frontal lobe.

The following signal streams were collected, which are grouped into five standard frequency bands (Noachtar et al., 1999):

α-Activity: Frequency range of 8-13 Hz, originating from the parietal and occipital lobes during states of awareness, calm, or rest. Blinking or thinking perturbs alpha waves, known as α-block.

β-Activity: Frequency range of 14-30 Hz, appearing during states of consciousness or wakefulness and sensory stimulation.

Ө-Activity: Frequency range of 4-7 Hz, occurring in the parietal and occipital lobes during psychological stress, interruption of awareness, and free time.

δ-Activity: Frequency range of 0.5-3 Hz, generated during deep sleep, or in cases of oxygen deprivation or unconsciousness.

γ-Activity: Frequency range of 31-50 Hz, related to directed attention and states of perception or emotional activity.

Different brain areas generate different EEG signals. The brain's electromagnetic activity was recorded using an international five-electrode system, in which electrodes are connected to obtain signals from different skull regions. This setup allowed for a comprehensive observation of EEG signals, making this technique practical and convenient.

To encourage the use of electrodes, a wireless EEG device was employed to inspect and interpret the EEG signals from the frontal lobes. The study's purpose was to ascertain students' concentration levels by using simple detection and classification methods. To enable the implementation of the proposed system, the students must familiarize themselves with the necessary instructions. Portable devices were used to collect the EEG signals, facilitating the ease of use and accessibility of this system. The correlation matrix of the dataset is shown in Figure 2.

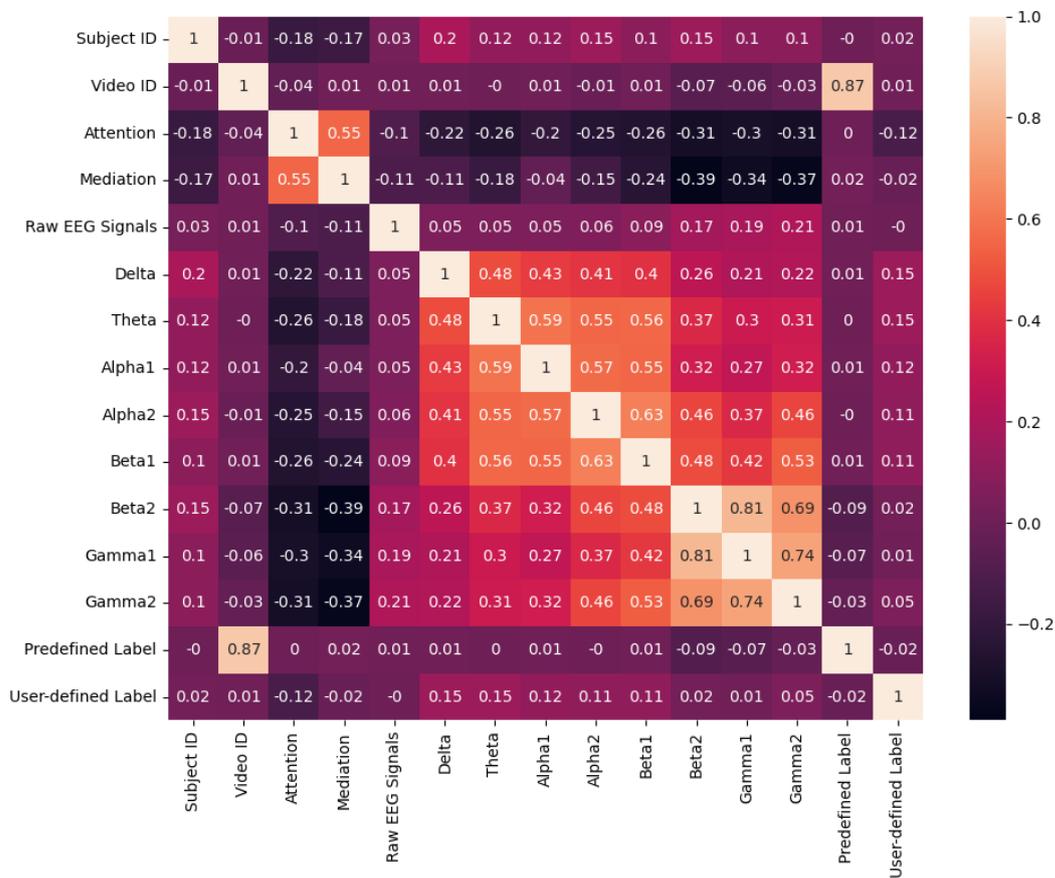

**Figure 2.** Correlation matrix

As the response vector is concerned, it contains the class variables of each row of the matrix. In a given data file, class variables are predefined labels. These variables only determine whether the student has learned the topic or not. Predefined and user-defined labels correspond to two states of mind. The former are used when students are expected to be confused, and the latter are labels attributed to students' own rating based on their experiences.

## 7. Analysis of results

This section discusses the observations made based on the classification metrics, such as precision, recall, the F1-score and the region enclosed by the receiver operating characteristic (ROC) curves.

### 7.1. Metrics-based performance

In this work, the analysed dataset was separated into two classes, one including the students who have learned and the other one the students who have not learned. The parameters of the confusion matrix were used to evaluate this binary classification.

**Table 1.** The names and formulas of the employed parameters

| Parameter | Formula |
|---|---|
| Precision | $\dfrac{TP}{TP + FP}$ |
| Recall | $\dfrac{TP}{TP + FN}$ |
| F1- score | $\dfrac{2 \cdot (Precision \cdot Recall)}{(Precision + Recall)}$ |
| Accuracy | $\dfrac{TP + TN}{TP + TN + FP + FN}$ |

### 7.2. Performance evaluation

The following classification models were employed in this study: Support Vector Machine (SVM), Naïve Bayes and Random Forest. Additionally, an ensemble approach was implemented by selecting the best performing classifier from the aforementioned models. The performance metrics indicated that the Naïve Bayes classifier achieved an accuracy of 89.50%, which was the lowest among the individual models. Referring to Table 2 attached below it can be seen that the ensemble model, however, demonstrated a superior performance with an accuracy of 99.92%. Interestingly, the Random Forest model achieved a perfect accuracy of 100% for the test data. Such a result might indicate overfitting, as it is uncommon for a model to achieve a perfect accuracy for real-world data. This observation suggests that the model may have overfitted the training data, indicating a lack of generalization capability for unfamiliar data. By contrast, ensemble methods can mitigate overfitting by capturing complementary patterns across different models. The obtained results clearly demonstrate that the combined model's accuracy surpasses that of any individual model. Furthermore, the ensemble model also excelled in other performance metrics, achieving a precision of 1.00, an F1-score of 1.00, and a recall of 1.00 as shown in Table 3 below. These scores indicate an exceptional performance across all of the evaluated metrics, underscoring the robustness and reliability of the ensemble approach in comparison with that of the individual models.

**Table 2.** Comparison of the employed classifiers with regard to their accuracy.

| Algorithm | Test Accuracy (%) |
|---|---|
| SVM | 99.72 |
| Random Forest | 100 |
| Naive Bayes | 89.50 |
| **Ensemble Model (With Bagging)** | **99.92** |

**Table 3.** Comparison of the employed models with regard to precision, recall and the F1-score

| Algorithm | Precision | | Recall | | F1-score | |
|---|---|---|---|---|---|---|
| | Class 0 | Class 1 | Class 0 | Class 1 | Class 0 | Class 1 |
| SVM | 1.00 | 1.00 | 1.00 | 1.00 | 1.00 | 1.00 |
| Random Forest | 1.00 | 1.00 | 1.00 | 1.00 | 1.00 | 1.00 |
| Naive Bayes | 0.96 | 0.84 | 0.83 | 0.96 | 0.89 | 0.90 |
| Ensemble Model | 1.00 | 1.00 | 1.00 | 1.00 | 1.00 | 1.00 |

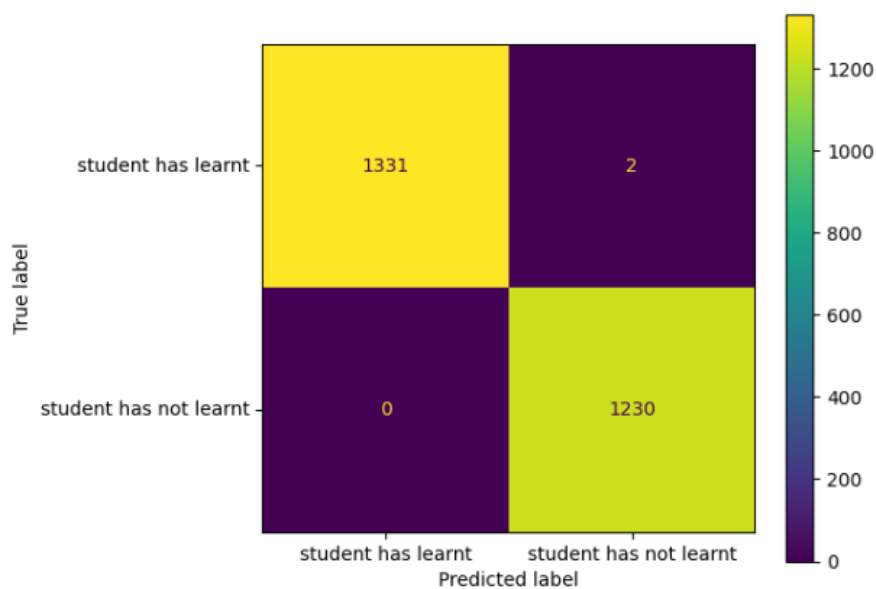

**Figure 3.** Confusion matrix for the ensemble model prediction performance

## 7.3. Receiver Operating Characteristic (ROC) curve analysis for the employed models

The classifiers utilised in this study exhibited varying performance levels across all categorization stages. These performance differences are illustrated through a Receiver Operating Characteristic (ROC) curve, which plots the true positive rate (TPR) against the false positive rate (FPR) for each threshold used to categorise data points. The performance of random prediction, random forest, and Naive Bayes classifiers were evaluated. The values for the area under the ROC curve (AUC) for these classifiers were 0.5, 1, and 0.978, respectively. Notably, the Random Forest classifier achieved the highest AUC value, which indicates a superior performance in balancing true positive and false positive rates. The performance of these models with regard to the ROC curve is depicted in Figure 4.

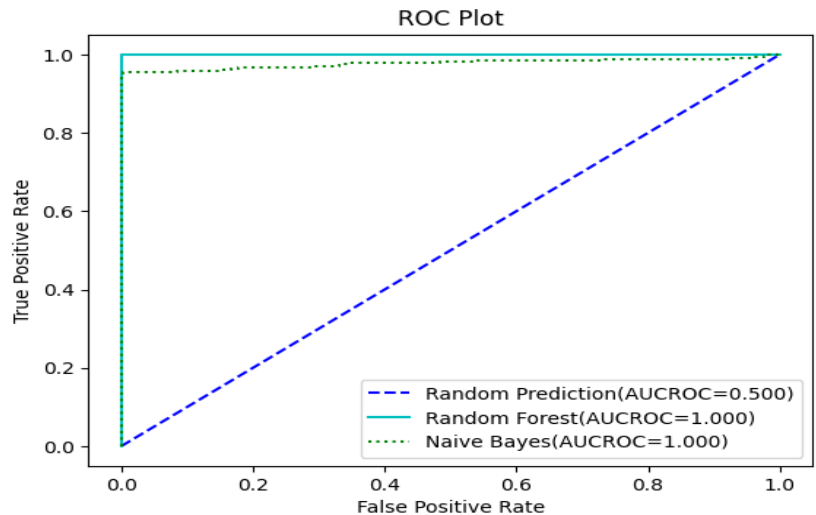

**Figure 4.** AUCROC plot for the employed classifiers

So, it can be stated that the Random Forest classifier forms an ensemble of all classifiers and that it has the highest threshold values, showing the highest performance rate of all.

### 7.4. Validation of proposed model

Predictive models were developed for the EEG dataset and then they were analysed using resampling techniques, with a view to providing quantifiable results. Finally, an analysis of statistical data was carried out to determine the most appropriate model. In this study, the average accuracy of classification was determined to be **99.76%** based on the Ensemble model being trained and tested using a stratified 10-fold cross-validation technique.

## 8. Conclusion

In today's digital world, various traditional techniques are used for teaching students. The old philosophy of face-to-face teaching is still thought to be the best of all. Teachers determine students' attentiveness within the classroom based on their facial expressions and gestures. However, it is difficult for teachers to determine the learning status of students when they are studying online or in a remote setting. Consequently, this study used quantified EEG data to analyse the student's concentration level so that teachers and students can be able to determine whether or not a subject is attentive. This will allow teachers to adjust the content of teaching, cultivating their learning attitudes, and enhance their focus while learning a topic. In this investigation, each characteristic was separately evaluated during the trials which were carried out. According to the obtained results, using 14 characteristics at once yielded the best classification accuracy, even though the accuracy of each classification method differs individually. Naive Bayes has variations and achieved a classification accuracy which was 10.50% lower than the average accuracy. According to Mohsin Ali Manzoor (Ali Shah *et al.* 2022) changes in accuracy suggest that Naive Bayes is the weakest model of all. It may be deduced that delta and beta waves are unrelated to the degree of attention exhibited by participants. Nonetheless, in this study an ensemble model and a precise sensing apparatus were used to identify EEG signals. The limitation of this approach lies in the fact that the dataset contains only 12000 data points which is quite a small number for training the model. This relatively small number can lead to the model becoming overfitted when using different sets of test data. Future research directions would include using large datasets to improve the classification results.

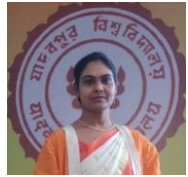

**Swati CHOWDHURI** is a Professor at the Institute of Engineering and Managemen in Kolkata, India. She obtained her Ph.D. and M. Tech. degrees at Jadavpur University in Kolkata and her Bachelor's degree at Burdwan University with Honours. Her research interests are in the field of MIMO communication, Wireless Networks, Deep Learning application in image processing, Machine Learning and Brain-Computer Interface. She published more than 40 works in international journals and conference proceedings. She has almost 17 years of teaching experience. She is a member of the Executive Committee of IEEE Women in Engineering (WIE), Kolkata Section and Life Member of IEI and ISTE.

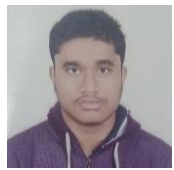

**Satadip SAHA** is a student pursuing a Bachelor's degree in Electrical and Electronics Engineering at the Institute of Engineering and Management, Kolkata. He is deeply interested in machine learning, robotics, electronics and network theory. He is especially passionate about neural networks and aims to study and bring about further development in this field. He is a member of the student chapter of the Institute of Engineering and Technology and has made significant contribution to its functioning.

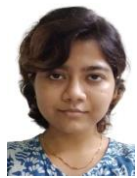

**Samadrita KARMAKAR** is currently pursuing her Bachelor's degree in Electrical and Electronics Engineering at the Institute of Engineering and Management, India. She has a keen interest in the functioning of the neural pathways of the brain and their replication using programmable languages. She is also passionate about the amalgamation of art and the field of electronics and has been working with digital media, focusing on the basic electronic elements that go into its creation.  She is currently working on honing her skills in diverse fields, aiming to become a multi-faceted academician. She is a member of the student chapter of the Institute of Engineering and Management.

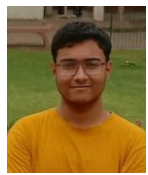

**Ankur CHANDA** is a final-year student pursuing a Bachelor's degree in Computer Science and Engineering with the specialization in Artificial Intelligence & Machine Learning at the Institute of Engineering and Management, India. His research interests are in the fields of Natural Language Processing for social good, Deep Learning, Machine Learning, Computer Vision, and Medical Imaging. He is a Research Intern at Jadavpur University, Kolkata, and has made significant contributions to their ongoing work.